%% file: fa_kpconv_paper.tex
\documentclass[conference]{IEEEtran}
\usepackage{cite}
\usepackage{hyperref}
\usepackage{booktabs}
\usepackage{multirow}
\usepackage{amsmath,amssymb,amsfonts}
\usepackage{algorithmic}
\usepackage{graphicx}
\usepackage{textcomp}
\usepackage{xcolor}
\def\BibTeX{{\rm B\kern-.05em{\sc i\kern-.025em b}\kern-.08em
    T\kern-.1667em\lower.7ex\hbox{E}\kern-.125emX}}
\begin{document}

\title{FA-KPConv: Introducing Euclidean Symmetries to KPConv via Frame Averaging}

\author{
	\IEEEauthorblockN{Ali Alawieh\IEEEauthorrefmark{1}\IEEEauthorrefmark{2} and Alexandru P. Condurache\IEEEauthorrefmark{1}\IEEEauthorrefmark{2}}\\
	\IEEEauthorblockA{\IEEEauthorrefmark{1}Robert Bosch GmbH - ADAS Systems, Software \& Services}
	\IEEEauthorblockA{\IEEEauthorrefmark{2}University of L\"ubeck - Institute for Signal Processing}
}

\maketitle

\begin{abstract}
	We present Frame-Averaging Kernel-Point Convolution (FA-KPConv), a neural network architecture built on top of the well-known KPConv, a widely adopted backbone for 3D point cloud analysis. Even though invariance and/or equivariance to Euclidean transformations are required for many common tasks, KPConv-based networks can only approximately achieve such properties when training on large datasets or with significant data augmentations. Using Frame Averaging, we allow to flexibly customize point cloud neural networks built with KPConv layers, by making them exactly invariant and/or equivariant to translations, rotations and/or reflections of the input point clouds. By simply wrapping around an existing KPConv-based network, FA-KPConv embeds geometrical prior knowledge into it while preserving the number of learnable parameters and not compromising any input information. We showcase the benefit of such an introduced bias for point cloud classification and point cloud registration, especially in challenging cases such as scarce training data or randomly rotated test data. We plan to open-source our implementation upon publication.
\end{abstract}


\section{Introduction}
Point cloud analysis has attracted a significant amount of research in recent years, especially given the widespread availability of point-cloud data. Being inherently unstructured, irregular and unordered, it is challenging to design neural network (NN) architectures to handle point clouds, and several approaches appeared over time. One of the well-known point cloud NN architectures is KPConv \cite{kpconv}, which has proven itself as a simple yet effective backbone, widely adopted until today for various point cloud analysis tasks \cite{gfnet,regtr,superpoint_matching,obpose,geotransformer}. Inspired by image convolution, KPConv defines point convolution using kernel points in Euclidean space, then applies the weights of those kernel points to the input points close to them.\\\\
\begin{figure}[!t]
	\centering
	\includegraphics[width=0.75\columnwidth]{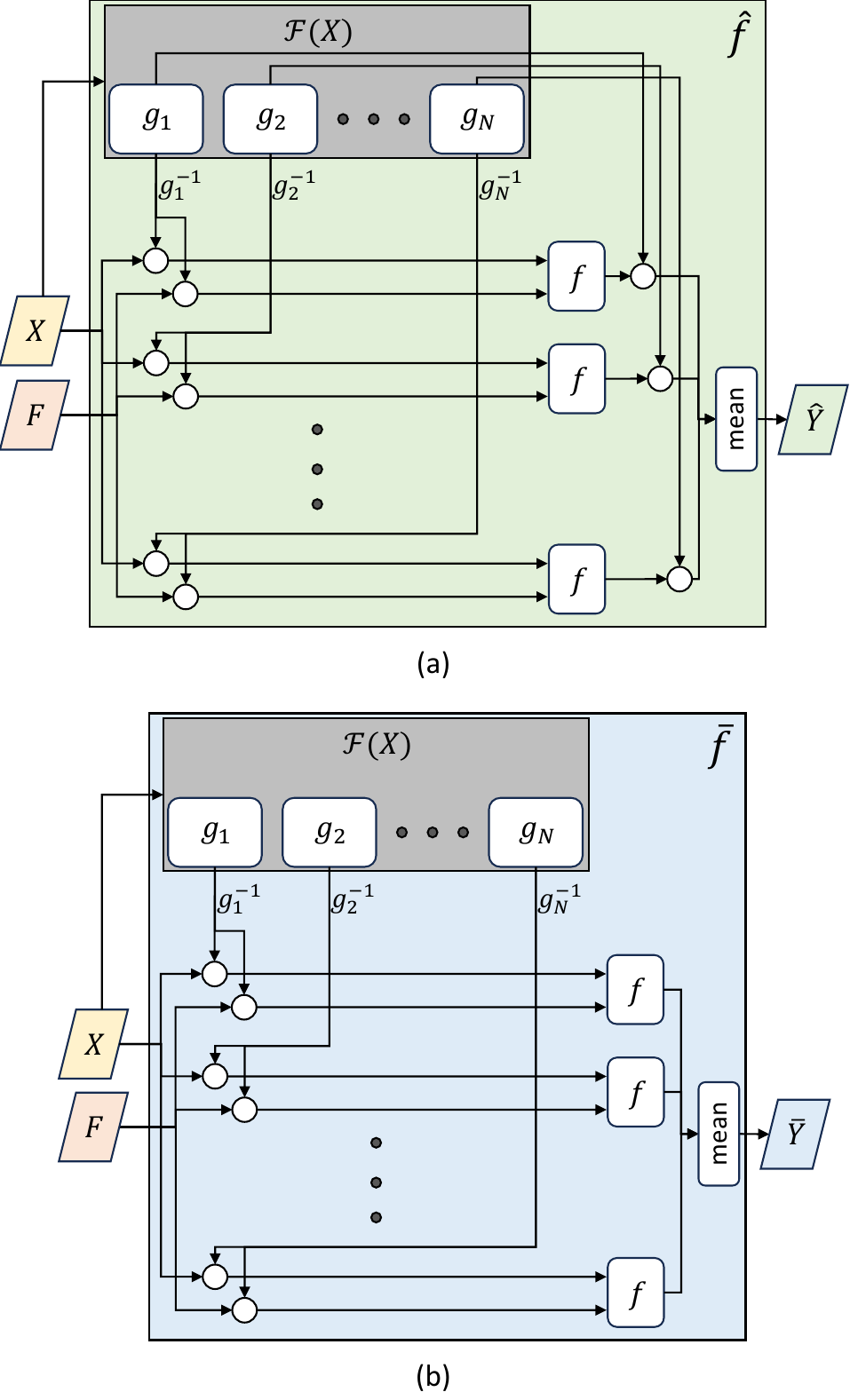}
	\caption{Diagrams illustrating how (a) the equivariant function $\hat{f}(\textbf{X},\textbf{F})$ and (b) the invariant function $\bar{f}(\textbf{X},\textbf{F})$ can be obtained from $f(\textbf{X},\textbf{F})$ via frame averaging, which is key for defining FA-KPConv. We use small circles to denote the action of a certain group element $g$, and denote $|\mathcal{F}(\textbf{X})|$ by $N$.}
	\label{flow_chart}
\end{figure}
Most of the known point cloud network architectures \cite{pcldl} were designed to ensure permutation invariance or permutation equivariance of the input data, which is crucial to deal with the inherent lack of order in the set of points constituting the point cloud. Furthermore, many tasks and use cases involve learning functions that are desired to be invariant or equivariant to other kinds of symmetries like rotations and translations. Being able to ensure such invariance or equivariance by design whenever needed is helpful, not only to embed and guarantee the desired properties, but also to improve the training sample efficiency and reduce the need for huge datasets or numerous data augmentations, while at the same time improving the explainability of the approach.\\\\
By integrating a recent elegant method called Frame Averaging \cite{fa} into KPConv, we enable making the popular KPConv backbone invariant or equivariant by design to Euclidean transformations of the input data. We do so without any overhead of additional trainable parameters, which results in better usage of the available model capacity of the network to learn other relevant cues from the training data.

\section{Related Work}
\textbf{Point cloud neural network architectures} were pioneered by PointNet \cite{pointnet}, which was the first architecture to directly operate on point clouds, i.e. without having to create a voxelized or a multi-view version of them. Essentially, it does so by passing each point of the point cloud through a shared multi-layer perceptron (MLP), then applying a permutation-invariant operation like max-pooling on the outputs to create a global feature vector for the point cloud.\\\\
To address the limitations of PointNet in capturing local structures, follow-up works tried to alleviate those limitations by recursively sampling, grouping neighborhoods and learning local features per group, thereby hierarchically going from local to global features. Works like PointNet++ \cite{pointnet++} and PointNeXt \cite{pointnext} employ MLPs per local neighborhood. Others like KPConv \cite{kpconv}, SpiderCNN \cite{spidercnn}, PointCNN \cite{pointcnn}, PAConv \cite{paconv} and PointConv \cite{pointconv} define various flavors of convolution operations, and apply them per local neighborhood. Further methods like DGCNN \cite{dgcnn}, ECC \cite{ecc} and AdaptConv \cite{adaptconv} model the point cloud as a graph, then define and apply graph convolution operations per local neighborhood.\\\\
A more recent line of research embraces the transformer architecture and adapts it for point clouds, thereby globally extracting features while inherently capturing local and non-local structures, with the help of attention mechanisms, at the expense though of quadratic computational complexity. PCT \cite{pct}, PT \cite{pt}, PTv2 \cite{ptv2}, PTv3 \cite{ptv3}, Point-BERT \cite{pointbert}, Point-MAE \cite{pointmae} and Point-M2AE \cite{pointm2ae} are example works along this line.\\\\
Even though KPConv \cite{kpconv} might not be the best performing among the mentioned architectures, yet it stands out for its intuitive design, inspired by image convolution, and for being a simple yet effective backbone still widely adopted in many recent works, and for various tasks \cite{gfnet,regtr,superpoint_matching,obpose,geotransformer}.\\\\
The main idea behind KPConv is to place kernel points in 3D space distributed around a center point, which is the point being convolved. Each kernel point is associated with a learnable weight matrix, and defines a region in space where it interacts with neighboring input points based on proximity. The kernel points are then placed around each query point at which the output is to be computed, and, using all input points lying in a radius neighborhood of the query point, the output at that point equals the sum of contributions over all pairs that can be formed using one input point from the neighborhood and one kernel point. The contribution of each such pair is the product of the weight matrix of the kernel point and the feature vector of the input point, weighted by the proximity between the kernel and the input point.\\

\textbf{In- or equivariance to Euclidean transformations} can either be approximately learned during training of a certain NN (e.g. via data augmentation), or introduced and enforced by design to its architecture.\\\\
What \textbf{approximate} methods are concerned, there are several promising approaches:
\begin{enumerate}
	\item One line of research focuses particularly on invariance and tries to approximately achieve it by estimating a transformation to align each input point cloud to a canonical space before feeding it into the NN. PointNet \cite{pointnet} for example employs what they call a joint alignment network, T-Net for short, which is a mini-version of PointNet itself aimed at regressing the affine transformation that aligns each input point cloud to a canonical space, thus achieving approximate $\mathrm{SE}(3)$ invariance, which is desired for classification tasks. T-Net was inspired by spatial transformer networks (STNs) \cite{stn} from the image domain, and was also adopted in several follow-up works after PointNet. It was shown however that T-Net based approaches usually require a significant amount of data augmentations to learn the approximate invariance \cite{gipointnet}.
	\item Another line of research focuses on equivariance for convolutional neural networks (CNNs), as in Group equivariant Convolutional Neural Networks (G-CNNs) \cite{gecnn} and Steerable CNNs \cite{stcnns} that can be applied on regular images, and in spherical CNNs \cite{spcnns} for spherical images. For point clouds, Equivariant Point Network (EPN) \cite{epn} combines group convolution with point convolution to achieve $\mathrm{SE}(3)$ equivariance. Even though those methods define group convolutions using group theory, which mathematically guarantees exact rotation equivariance, they are in practice applied on a finite group of rotations such as icosahedral grids, not over the intractable infinite continuous group of rotations. That's why they can still be considered as approximate methods.\\
\end{enumerate}
What \textbf{exact} methods are concerned, there are also several promising approaches:
\begin{enumerate}
	\item A first line of research was pioneered by the seminal work entitled Tensor Field Networks (TFNs) \cite{tfn}. By relying on irreducible representations of the rotation group, TFN designs convolution layers that are $\mathrm{SE}(3)$-equivariant by ensuring that both feature maps and learned filters transform consistently under $\mathrm{SE}(3)$ transformations. Another known work in that direction is $\mathrm{SE}(3)$-Transformer \cite{se3tr} which extends TFN from convolution to transformer layers, with equivariant self-interaction and message passing / attention mechanisms. It is worth noting that this set of methods is mathematically quite involved and complex.
	\item Another line of research achieves exact invariance by hand-crafting, from the input point clouds, values that are invariant by design. Example invariant values are distances, angles, dot and cross products between vectors defined from the points lying in a certain neighborhood, or from the normal vectors at those points, as done in \cite{clusternet,riconv,ppfnet,srinet}. Another way to define invariant values is by first constructing one or more rotation-equivariant local reference frame(s) (LRF) like the Darboux frame \cite{darboux}, then projecting relative point coordinates to the LRF(s) to render them $\mathrm{SE}(3)$-invariant.
	\item A third line of research achieves exact $\mathrm{SE}(3)$ in- or equivariance by defining or modifying NN layers to make them as such. One prominent example is Equivariant Graph Neural Networks (EGNN) \cite{egnn}, in which the message passing operations of the graph neural network are designed to update both node features and relative coordinates between the nodes in a way that preserves equivariance. Another seminal work is Vector Neurons (VN) \cite{vn}, which represents features as 3-dimensional vectors rather than scalars, and maintains those representations throughout the network by building equivariance into common NN layers (like linear, non-linear, pooling and normalization layers).
	\item A unique approach enjoying the advantages and avoiding the disadvantages of the other approaches is called Frame Averaging \cite{fa}. It does not require designing any dedicated NN layers, and is applicable to any available point cloud NN via simple plug-and-play, without having to modify its layers, or to add any parameters to be learned to it. It is also exact, not approximate, since it is based on group theory, and avoids having to hand-craft input features and the potential information loss associated to that. The idea behind it is to simply replace group averaging, which is intractable for an infinite group like $\mathrm{SE}(3)$ with frame averaging, where a frame is a carefully selected and input-dependent small subset of that group, on which averaging is shown to be exactly equivalent to averaging over the whole group.\\
\end{enumerate}
Further insights on methods for embedding geometrical prior knowledge into NNs are surveyed in \cite{rathsurvey}, a recent survey on that field.\\\\
Our contribution in this work consists mainly of the following:
\begin{itemize}
	\item We create FA-KPConv by integrating \textbf{F}rame \textbf{A}veraging with the widely-used KPConv backbone, thereby enabling users to easily embed geometrical prior knowledge into that popular architecture. More specifically, we enable the introduction of exact invariance or equivariance to Euclidean transformations of the input point cloud.
	\item We showcase the performance boost, the improved learning capacity and the sample efficiency benefits of FA-KPConv on point cloud classification and registration tasks, especially for geometrically transformed datasets and in the low-data regime.\\
\end{itemize}

\section{Methodology}
Our approach integrates Frame Averaging \cite{fa} and KPConv \cite{kpconv} into FA-KPConv. Details of the approach are provided in the following sections.

\subsection{Frame Averaging for Point Cloud Networks}\label{fa_section}
We start by recalling some basics related to equivariance, invariance and group theory \cite{rathsurvey}.\\\\
A function $f: X \rightarrow Y$ is equivariant with respect to a transformation group $G$ if, for any input $\textbf{X} \in X$ and transformation $g \in G$, transforming $\textbf{X}$ by $g$ then feeding $g \cdot \textbf{X}$ to $f$ gives the same result as feeding $\textbf{X}$ to $f$ then transforming $f(\textbf{X})$ by $g'$, where $g'$ is the transformation corresponding to $g$ in the output space $Y$, i.e.:
\begin{equation} \label{equiv_def}
f(g \cdot \textbf{X})=g' \cdot f(\textbf{X}) \qquad \forall \textbf{X} \in X, \; \forall g \in G
\end{equation}
Invariance on the other hand is a special case of equivariance, where the transformation in output space is the identity, i.e. the output is unaffected by transformations $g \in G$:
\begin{equation} \label{inv_def}
f(g \cdot \textbf{X})=f(\textbf{X}) \qquad \forall \textbf{X} \in X, \; \forall g \in G
\end{equation}
\newline
Moreover, an arbitrary function $f: X \rightarrow Y$ can be made equivariant or invariant to a certain transformation group $G$ by \textit{symmetrization}, that is averaging over the group \cite{rathsurvey}, i.e. starting from $f$, we can obtain an equivariant function $\hat{f}$ or an invariant function $\bar{f}$ as such:
\begin{equation} \label{equiv_ga}
\hat{f}(\textbf{X})=\frac{1}{|G|}\sum_{g \in G}{g \cdot f(g^{-1} \cdot \textbf{X})}
\end{equation}
\begin{equation} \label{inv_ga}
\bar{f}(\textbf{X})=\frac{1}{|G|}\sum_{g \in G}{f(g^{-1} \cdot \textbf{X})}
\end{equation}
\newline
\textit{Symmetrization} becomes intractable when the cardinality of $G$ is large or infinite, as is the case for the continuous group of 3D translations or rotations. This intractability made many works resort to approximate in- or equivariance by averaging over a subset of the transformation group \cite{gecnn,stcnns,epn}.\\\\
Frame Averaging \cite{fa} elegantly avoids both issues by averaging over a small input-dependent subset of transformations, designed and chosen in a way that guarantees exact equivalence between averaging over it and the intractable averaging over the whole group. That subset is called a frame $\mathcal{F}(\textbf{X})$ , and equations \ref{equiv_ga} \& \ref{inv_ga} become:
\begin{equation} \label{equiv_fa}
\hat{f}(\textbf{X})=\frac{1}{|\mathcal{F}(\textbf{X})|}\sum_{g \in \mathcal{F}(\textbf{X})}{g \cdot f(g^{-1} \cdot \textbf{X})}
\end{equation}
\begin{equation} \label{inv_fa}
\bar{f}(\textbf{X})=\frac{1}{|\mathcal{F}(\textbf{X})|}\sum_{g \in \mathcal{F}(\textbf{X})}{f(g^{-1} \cdot \textbf{X})}
\end{equation}
\begin{table*}[h!]
	\begin{center}
		\caption{Frames for groups of Euclidean motions in $\mathbb{R}^d$.\protect\linebreak$\mathrm{O}(d)$, $\mathrm{SO}(d)$, $\mathrm{E}(d)$ and $\mathrm{SE}(d)$ denote respectively the orthogonal, special orthogonal, Euclidean and special Euclidean groups.}
		\label{frames}
		\begin{tabular}{c|c|c|c|c}
			\toprule
			\multicolumn{3}{c|}{\textbf{Group}} & \multicolumn{2}{c}{\textbf{Frame}} \\
			Name & Elements & Representation for $\textbf{X} \in \mathbb{R}^{n \times d}$ & $\mathcal{F}(\textbf{X})$ & $|\mathcal{F}(\textbf{X})|$ \\
			\midrule
			Translations & $\textbf{t} \in \mathrm{T}(d)=\mathbb{R}^d$ & $\textbf{X}+\textbf{1}\textbf{t}^T$ & $\{\textbf{c}\}$ & 1\\
			Rotations & $\textbf{R} \in \mathrm{SO}(d)$ & $\textbf{XR}^T$ & $\{\textbf{Q}\mid\textbf{C}=\textbf{Q}\Lambda\textbf{Q}^T,|\textbf{Q}|=1\}$ & $2^{d-1}$\\
			Rotations \& Reflections & $\textbf{R} \in \mathrm{O}(d)$ & $\textbf{XR}^T$ & $\{\textbf{Q}\mid\textbf{C}=\textbf{Q}\Lambda\textbf{Q}^T\}$ & $2^d$ \\
			Translations \& Rotations & $(\textbf{R},\textbf{t}) \in \mathrm{SE}(d)$ & $\textbf{XR}^T+\textbf{1}\textbf{t}^T$ & $\{(\textbf{Q},\textbf{c})\mid\textbf{C}=\textbf{Q}\Lambda\textbf{Q}^T,|\textbf{Q}|=1\}$ & $2^{d-1}$\\
			Translations \& Rotations \& Reflections & $(\textbf{R},\textbf{t}) \in \mathrm{E}(d)$ & $\textbf{XR}^T+\textbf{1}\textbf{t}^T$ & $\{(\textbf{Q},\textbf{c})\mid\textbf{C}=\textbf{Q}\Lambda\textbf{Q}^T\}$ & $2^d$ \\
			\bottomrule
		\end{tabular}
	\end{center}
\end{table*}
Such a frame might not exist though for all possible functions $f: X \rightarrow Y$ and transformation groups $G$. Reference \cite{fa} proves the existence of such frames for any function that operates on point clouds (thus including point cloud NNs) and for groups of Euclidean motions in $\mathbb{R}^d$. Formally, for a point cloud $\textbf{X} \in \mathbb{R}^{n \times d}$, $n$ being the number of points in the point cloud, and using $\textbf{1}$ to denote a vector of ones $\in \mathbb{R}^n$, we first compute the point cloud centroid $\textbf{c}$ and the covariance matrix $\textbf{C}$ as such:
\begin{equation} \label{centroid}
\textbf{c}=\frac{1}{n}\textbf{X}^T\textbf{1}
\end{equation}
\begin{equation} \label{covariance}
\textbf{C}=(\textbf{X}-\textbf{1}\textbf{c}^T)^T(\textbf{X}-\textbf{1}\textbf{c}^T)
\end{equation}
We then obtain the matrix $\textbf{Q}$ by normalized Eigendecomposition of $\textbf{C}$:
\begin{equation} \label{eigendecomp}
\textbf{C}=\textbf{Q}\Lambda\textbf{Q}^T
\end{equation}
Note that the columns of $\textbf{Q}$ consist of unit-length eigenvectors, and are unique up to a sign change. Thus, we have $2^d$ possible solutions for $\textbf{Q}$ in equation (\ref{eigendecomp}).\\\\
Using $\textbf{c}$ and $\textbf{Q}$, the frames for the groups of Euclidean motions in $\mathbb{R}^d$ are listed in Table \ref{frames}. Further details and proofs can be found in \cite{fa}.
\subsection{KPConv}
Starting from a point cloud whose coordinates are represented as $\textbf{X} \in \mathbb{R}^{n \times d}$ and features as $\textbf{F}^{in} \in \mathbb{R}^{n \times c_{in}}$, KPConv \cite{kpconv} computes the output feature vector $\textbf{f}^{out} \in \mathbb{R}^{c_{out}}$ at a query point $\textbf{q} \in \mathbb{R}^{d}$ using point convolution by a kernel $h$ as such:
\begin{equation} \label{kpconv_1}
\textbf{f}^{out}=\sum_{i\in \mathcal{N}_\textbf{q}}{h(\textbf{X}_i-\textbf{q})\textbf{F}_{i}^{in}}
\end{equation}
where $\mathcal{N}_\textbf{q}$ is the radius neighborhood of $\textbf{q}$ from $\textbf{X}$ for a certain radius $r$, i.e. $\mathcal{N}_\textbf{q}=\{j\in \{0,1,\dots,n-1\}\mid\|\textbf{X}_j-\textbf{q}\|\leq r\}$.\\\\
To define $h$, we first define $\{(\tilde{\textbf{x}}_k,\textbf{W}_k)\in \mathbb{R}^d\times\mathbb{R}^{c_{in}\times c_{out}}\mid k<K,\|\tilde{\textbf{x}}_k\|\leq r\}$ as the set of $K$ kernel points with their associated weight matrices. Then,
\begin{equation} \label{kpconv_2}
h(\textbf{y}_i)=\sum_{k<K}{l(\textbf{y}_i,\tilde{\textbf{x}}_k)\textbf{W}_k}
\end{equation}
\begin{equation} \label{kpconv_3}
l(\textbf{y}_i,\tilde{\textbf{x}}_k)=\max\left(0,1-\dfrac{\|\textbf{y}_i-\tilde{\textbf{x}}_k\|}{\sigma}\right)
\end{equation}
with $\sigma$ being a hyperparameter reflecting the influence distance of the kernel points.\\\\
KPConv comes in two flavors: either \textbf{rigid}, in which the kernel point positions $\{\tilde{\textbf{x}}_k\}$ are determined beforehand then fixed during training, or \textbf{deformable}, in which they are learned via backpropagation during training.\\
By stacking KPConv layers together with other common layers like leaky ReLU, pooling, upsampling and batch normalization layers, \cite{kpconv} goes on to define KPConv-based neural network architectures, namely KP-CNN for point cloud classification, and KP-FCNN for point cloud segmentation.

\subsection{FA-KPConv}
A KPConv-based NN architecture can be represented as:
\begin{equation} \label{kpconv_nn}
\textbf{Y}=f(\textbf{X},\textbf{F}^{in})
\end{equation}
with $\textbf{X} \in \mathbb{R}^{n \times d}$ and $\textbf{F}^{in} \in \mathbb{R}^{n \times c_{in}}$ as before. $f$ in this equation can either be a single rigid/deformable KPConv layer, or more generally any architecture built using an arbitrary number of such layers together with other common NN layers, as is the case in KP-CNN and KP-FCNN. The dimensionality and shape of $\textbf{Y}$, the output, depend on the details of the architecture and the task to be solved.\\\\
In order to make $f$ invariant or equivariant to any Euclidean transformation group, we require $c_{in}$ to be a multiple of $d$, i.e. $c_{in}=k\cdot d$, to make it possible to temporarily reshape $\textbf{F}^{in}$ from $n \times c_{in}$ to $nk\times d$, then apply the transformation as listed in the \textit{Representation} column in Table \ref{frames}, and then reshape the output back to the original shape.\\\\
To make $f$ equivariant, we also require the last dimension of $\textbf{Y}$ to be a multiple of $d$, for similar reasons.\\\\
We first consider the case in which the input feature at each point is set to be equal to the coordinates of that point, as is commonly done in the literature. Formally, $\textbf{F}^{in}=\textbf{X}$, and $f$ becomes only a function of $\textbf{X}$, as in section \ref{fa_section}.\\\\
For that case, we can simply use the frame definition from Table \ref{frames} together with equation \ref{equiv_fa} or \ref{inv_fa} to make $f$ respectively equivariant or invariant to the transformation group corresponding to that frame. We can also do function composition of \ref{equiv_fa} and \ref{inv_fa} to achieve a mixture of invariance and equivariance on non-intersecting groups from that table. One example use case is to jointly achieve $\mathrm{T}(d)$-invariance with $\mathrm{O}(d)$-equivariance.\\\\
Even though the desired exact invariance and/or equivariance properties are theoretically proven in \cite{fa} to hold for that specific case in which $f$ is only a function of $\textbf{X}$, it turns out that they also exactly hold as well when $f$ is a function of both $\textbf{X}$ and $\textbf{F}^{in}$. We verified this via extensive testing.\\\\
Formally, an equivariant function $\hat{f}$ or an invariant function $\bar{f}$ can be defined out of $f(\textbf{X},\textbf{F}^{in})$ as such:
\begin{equation} \label{equiv_fa_}
\hat{f}(\textbf{X},\textbf{F}^{in})=\frac{1}{|\mathcal{F}(\textbf{X})|}\sum_{g \in \mathcal{F}(\textbf{X})}{g \cdot f(g^{-1} \cdot \textbf{X},g^{-1} \cdot \textbf{F}^{in})}
\end{equation}
\begin{equation} \label{inv_fa_}
\bar{f}(\textbf{X},\textbf{F}^{in})=\frac{1}{|\mathcal{F}(\textbf{X})|}\sum_{g \in \mathcal{F}(\textbf{X})}{f(g^{-1} \cdot \textbf{X},g^{-1} \cdot \textbf{F}^{in})}
\end{equation}
Check Figure \ref{flow_chart} for a visual illustration of those equations.\\\\
Any KPConv-based NN architecture following \ref{kpconv_nn} becomes an FA-KPConv architecture by using \ref{equiv_fa_} or \ref{inv_fa_} or a composition of both (assuming that $c_{in}$ and the last dimension of $\textbf{Y}$ are as required).\\\\
Note that the achieved equivariance or invariance is for the case in which the same transformation $g$ is applied to both $\textbf{X}$ and $\textbf{F}^{in}$, i.e. $\hat{f}$ and $\bar{f}$ satisfy the following equations:
\begin{equation} \label{equiv_def_}
\hat{f}(g \cdot \textbf{X},g \cdot \textbf{F}^{in})=g' \cdot \hat{f}(\textbf{X},\textbf{F}^{in})
\end{equation}
\begin{equation} \label{inv_def_}
\bar{f}(g \cdot \textbf{X},g \cdot \textbf{F}^{in})=\bar{f}(\textbf{X},\textbf{F}^{in})
\end{equation}
\newline
Since a FA-KPConv architecture simply wraps around its KPConv counterpart, it preserves the same number of trainable parameters. However, as can be seen in equations \ref{equiv_fa_} and \ref{inv_fa_}, the amount of memory consumed and that of computations involved do get multiplied by a factor roughly equal to $|\mathcal{F}(\textbf{X})|$, which is greater than or equal to 1 (cf. table \ref{frames}).
\section{Experiments}
We focus in our experiments on benchmarking KPConv vs FA-KPConv models on two popular 3D tasks: shape classification and point cloud registration.\\\\
For each task/benchmark, we run several experiments, and in each experiment, we train both KPConv and FA-KPConv models from scratch until convergence, while using the same training and testing configurations for both.\\\\
On the training side, each experiment uses a certain portion of the training data. This helps showcasing the training sample efficiency of FA-KPConv. On the testing side, we always use the whole test data available in each benchmark for testing, but we once test on it in its original form, and once on a rotated version of it, in which each sample is randomly rotated.
\subsection{3D Shape Classification}
3D shape classification is a task in which $\mathrm{\mathbf{E}}\mathbf{(3)}$\textbf{-invariance} is desired, since translations, rotations and reflections of a point cloud do not change its underlying class.\\\\
We use the ModelNet40 benchmark \cite{modelnet}, which contains 12,311 meshed 3D CAD models (9843 for training and 2468 for testing) from 40 categories.\\\\
We choose the KP-CNN architecture from \cite{kpconv} with rigid KPConv layers for that task since it showed the best performance on it. The original KP-CNN assigns to each input point a constant feature equal to 1, i.e. $c_{in}=1$, which is not a multiple of 3. We therefore define from it what we call baseline KP-CNN by simply setting $c_{in}=3$ and assigning a vector $\textbf{1} \in\mathbb{R}^{3}$ to each input point. FA-KPCNN is obtained from the baseline KP-CNN using equation \ref{inv_fa_} and $\mathcal{F}(\textbf{X})=\{(\textbf{Q},\textbf{c})\mid\textbf{C}=\textbf{Q}\Lambda\textbf{Q}^T\}$ from Table \ref{frames}.\\\\
We follow the same training and testing procedure as in the shape classification experiments in \cite{kpconv} for both models, and use Overall accuracy (OA) as testing score. Results can be found in Table \ref{modelnet_results}.
\begin{table}[t!]
	\begin{center}
		\caption{Testing OA (\%) of the baseline KP-FCNN vs FA-KPFCNN on the original and the rotated testing data of ModelNet40, when trained on different portions of the training data.}
		\label{modelnet_results}
		\begin{tabular}{c|c|c|c|c}
			\toprule
			\multicolumn{5}{c}{Overall Accuracy OA (\%) $\uparrow$} \\
			\midrule
			\# Training & \multicolumn{2}{c|}{KP-CNN (baseline)} & \multicolumn{2}{c}{FA-KPCNN (ours)}\\
			Samples & original & rotated & original & rotated\\
			\midrule
			500  & $\textbf{8.5}$ & $3.4$ & $8.0$ & $\textbf{7.3}$\\
			2000 & $\textbf{83.3}$ & $24.4$ & $74.4$ & $\textbf{74.5}$\\
			5000 & $\textbf{87.4}$ & $34.0$ & $83.2$ & $\textbf{82.9}$\\
			9843 & $\textbf{90.4}$ & $44.6$ & $87.1$ & $\textbf{87.0}$\\
			\bottomrule
		\end{tabular}
	\end{center}
\end{table}
\begin{table}[t!]
	\begin{center}
		\caption{Testing metrics of the baseline GeoTransformer vs FA-GeoTransformer on the original and the rotated testing data of 3DMatch/3DLoMatch, when trained on different portions of the training data.}
		\label{3dmatch_results}
		\begin{tabular}{ccccccc}
			\toprule
			\# Training & GeoTransformer & IR$\uparrow$ & FMR$\uparrow$ & RR$\uparrow$ & RRE$\downarrow$ & RTE$\downarrow$ \\
			Samples & Model & (\%) & (\%) & (\%) & ($^{\circ}$) & (cm)\\
			\midrule
			\multicolumn{7}{c}{3DMatch (original)} \\
			\midrule
			\multirow{2}*{1k} & baseline & 51.6 & \textbf{96.8} & \textbf{86.4} & 6.7 & 2.07\\
			& ours & \textbf{54.7} & 96.5 & 86.3 & \textbf{6.5} & \textbf{1.98}\\
			\multirow{2}*{5k} & baseline & 66.5 & \textbf{97.8} & 89.1 & 6.5 & 1.95\\
			& ours & \textbf{69.0} & 97.4 & \textbf{90.2} & \textbf{6.1} & \textbf{1.85}\\
			\midrule
			\multicolumn{7}{c}{3DLoMatch (original)} \\
			\midrule
			\multirow{2}*{1k} & baseline & 21.8 & 73.9 & 56.4 & 9.4 & 3.38\\
			& ours & \textbf{24.6} & \textbf{76.4} & \textbf{60.9} & \textbf{8.9} & \textbf{3.13}\\
			\multirow{2}*{5k} & baseline & 37.0 & 83.8 & 67.2 & \textbf{9.1} & \textbf{3.08}\\
			& ours & \textbf{39.6} & \textbf{85.9} & \textbf{71.4} & \textbf{9.1} & 3.09\\
			\midrule
			\multicolumn{7}{c}{3DMatch (rotated)} \\
			\midrule
			\multirow{2}*{1k} & baseline & 48.7 & \textbf{96.9} & 84.0 & 6.5 & 2.03\\
			& ours & \textbf{53.5} & 96.5 & \textbf{85.9} & \textbf{6.2} & \textbf{1.95}\\
			\multirow{2}*{5k} & baseline & 64.1 & \textbf{98.3} & 88.1 & 6.2 & 1.88\\
			& ours & \textbf{68.2} & 97.4 & \textbf{89.6} & \textbf{5.9} & \textbf{1.82}\\
			\midrule
			\multicolumn{7}{c}{3DLoMatch (rotated)} \\
			\midrule
			\multirow{2}*{1k} & baseline & 19.8 & 70.5 & 49.3 & 9.1 & 3.42\\
			& ours & \textbf{23.7} & \textbf{75.3} & \textbf{56.9} & \textbf{8.2} & \textbf{3.03}\\
			\multirow{2}*{5k} & baseline & 34.2 & 82.3 & 60.8 & 8.3 & 2.93\\
			& ours & \textbf{38.9} & \textbf{86.2} & \textbf{65.5} & \textbf{8.2} & \textbf{2.85}\\
			\bottomrule
		\end{tabular}
	\end{center}
\end{table}
\begin{table}[t!]
	\begin{center}
		\caption{Relative improvement in testing metrics of FA-GeoTransformer wrt. the baseline GeoTransformer on the original and the rotated testing data of 3DMatch/3DLoMatch, when trained on different portions of the training data.}
		\label{3dmatch_results_delta}
		\begin{tabular}{cccccc}
			\toprule
			\# Training & $\Delta$IR & $\Delta$FMR & $\Delta$RR & $\Delta$RRE & $\Delta$RTE \\
			Samples & (\%) & (\%) & (\%) & (\%) & (\%)\\
			\midrule
			\multicolumn{6}{c}{3DMatch (original)} \\
			\midrule
			1k & +6.0 & -0.3 & -0.1 & +4.8 & +4.3\\
			5k & +3.8 & -0.4 & +1.2 & +6.2 & +5.1\\
			\midrule
			\multicolumn{6}{c}{3DLoMatch (original)} \\
			\midrule
			1k & +12.8 & +3.4 & +8.0 & +5.3 & +7.4\\
			5k & +7.0 & +2.5 & +6.3 & 0.0 & -0.3\\
			\midrule
			\multicolumn{6}{c}{3DMatch (rotated)} \\
			\midrule
			1k & +9.9 & -0.4 & +2.3 & +4.6 & +3.9\\
			5k & +6.4 & -0.9 & +1.7 & +3.0 & +3.2\\
			\midrule
			\multicolumn{6}{c}{3DLoMatch (rotated)} \\
			\midrule
			1k & +19.7 & +6.8 & +15.4 & +9.9 & +11.4\\
			5k & +13.7 & +4.7 & +7.7 & +1.2 & +2.7\\
			\bottomrule
		\end{tabular}
	\end{center}
\end{table}
\subsection{Point Cloud Registration}
Point cloud registration is the task of determining the relative Euclidean transformation that aligns two input point clouds. We base our experiments on one of the popular and recent NN architectures, called GeoTransformer \cite{geotransformer}, which uses a KPConv-based feature extractor backbone in the first stage of its architecture, thereby generating features for estimating correspondences between the two input point clouds, then estimating the desired relative transformation using those correspondences.\\\\
Correspondence estimation, which is key to enable an accurate and robust registration, is a task in which $\mathrm{\mathbf{SE}}\mathbf{(3)}$\textbf{-invariance} is desired, since translations and rotations of either of the two input point clouds do not change what points correspond to each other in the underlying real-world environment.\\\\
We use the 3DMatch benchmark \cite{3dmatch}, which consists of pairs of point clouds with the ground-truth transformation that aligns each pair. The point clouds are RBG-D scans from indoor environments, and the benchmark is split into 3DMatch and 3DLoMatch, the former of which consists of pairs in which the overlap between the two point clouds of each pair is higher than 30\%, and the latter is the more challenging one, with the overlap being only 10-30\%.\\\\
We use the same architecture from \cite{geotransformer} along with its KPConv backbone. As done for shape classification, we define the baseline GeoTransformer from the original GeoTransformer by assigning a vector $\textbf{1} \in\mathbb{R}^{3}$ to each input point, instead of a constant feature equal to 1, i.e. we make $c_{in}=3$ instead of 1. FA-GeoTransformer is obtained from the baseline GeoTransformer using equation \ref{inv_fa_} and $\mathcal{F}(\textbf{X})=\{(\textbf{Q},\textbf{c})\mid\textbf{C}=\textbf{Q}\Lambda\textbf{Q}^T,|\textbf{Q}|=1\}$ from Table \ref{frames}, both of which are applied on the KPConv backbone inside GeoTransformer.\\\\
We follow the same training and testing procedure as in \cite{geotransformer} for both models, and use the same metrics, i.e. Inlier Ratio (IR), Feature Matching Recall (FMR), Registration Recall (RR), Relative Translation Error (RTE), and Relative Rotation Error (RRE) for evaluation. Note that IR and FMR measure the quality of the correspondence estimation, while RR, RRE and RTE measure the quality of the transformation estimation.\\\\
Results can be found in Table \ref{3dmatch_results}, and, for convenience, Table \ref{3dmatch_results_delta} illustrates the improvements of our model vs the baseline for the different setups.
\begin{figure}[!t]
	\centering
	\includegraphics[width=\columnwidth]{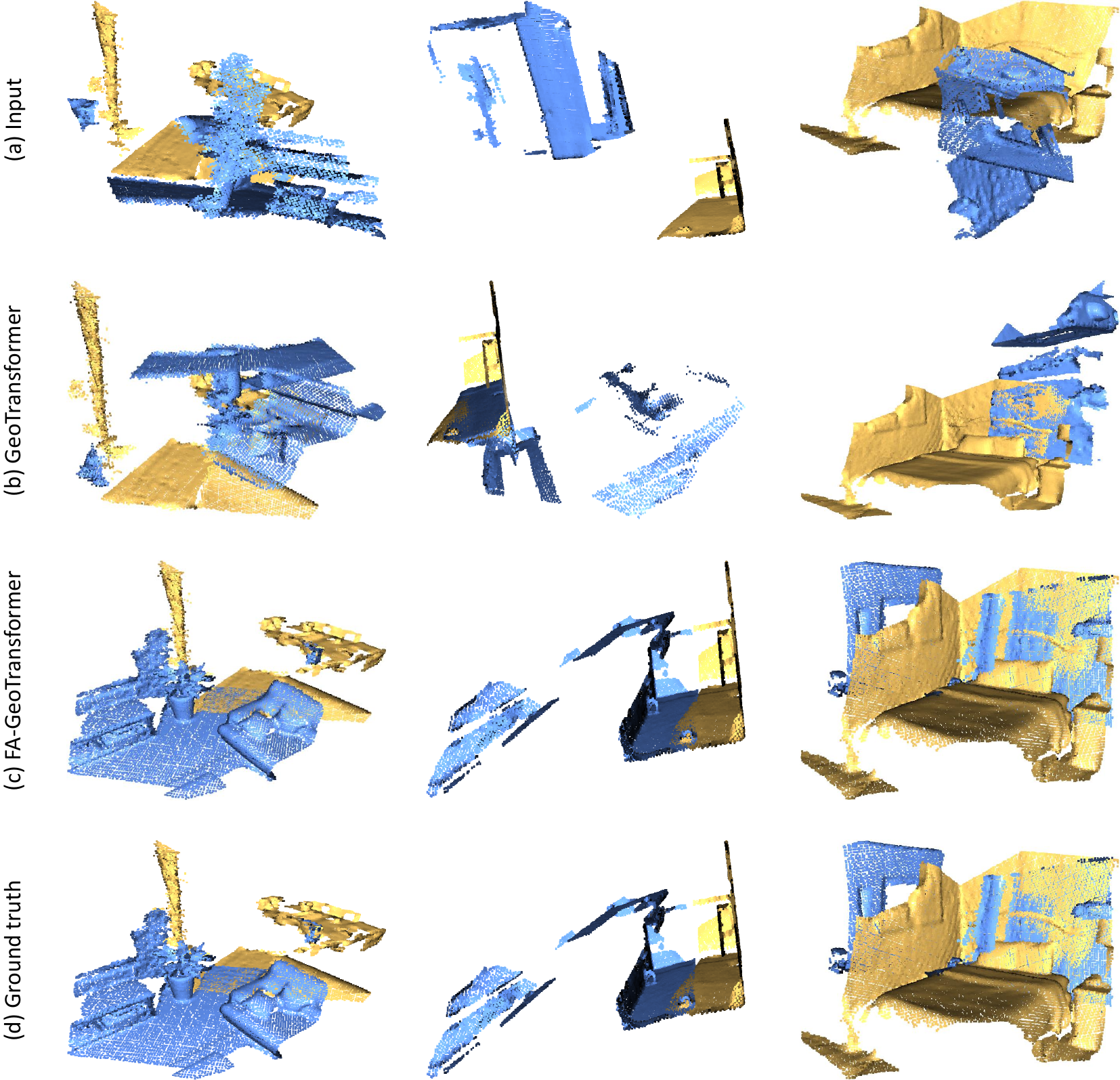}
	\caption{Qualitative comparison of registration results on rotated 3DLoMatch with models trained on 1k samples.}
	\label{qualitative_results}
\end{figure}
\section{Analysis \& Discussion}\label{analysis}
Looking at the results tables \ref{modelnet_results}, \ref{3dmatch_results} and \ref{3dmatch_results_delta}, we first note the fact that for each benchmark, the number of trainable model parameters for the KPConv and that of its FA-KPConv counterpart are exactly the same. However, since the FA-KPConv models embed, via frame averaging, geometrical prior knowledge, they actually can allocate more capacity to learning other relevant invariances. Nevertheless, we observe that in some settings, FA-KPConv models do not improve over the baseline. These settings are characterized by the fact that the embedded invariances are not needed in the respective problem setup. We conjecture that in this case, the model spends capacity on learning to ignore the invariances imposed by design through FA, which leads to poor performance.\\\\
This can be clearly observed in the results of the ModelNet40 experiments that are done on the original testing data. In those experiments, both the original training and testing datasets only contain shapes that are aligned in orientation to a canonical frame, thus rendering the $\mathrm{E}(3)$-invariance achieved by FA-KPCNN useless, and even harmful model complexity.\\\\
In fact, we notice from the experiments, that the advantage of FA-KPConv models over KPConv models becomes, in most cases, more visible the more challenging the problem at hand is:
\begin{enumerate}
	\item \textbf{Randomly rotated testing data}: As we would naturally expect, the performance of FA-KPConv models, which are specifically designed to be invariant (among other things) to rotations is better than that of KPConv models on rotated testing data. And the degradation in performance that KPConv models suffer from when switching from original to rotated data is much more significant, than that suffered by FA-KPConv models, if any. From tables \ref{modelnet_results}, \ref{3dmatch_results} and \ref{3dmatch_results_delta}, one can clearly observe that by comparing the performance of the baselines and that of our models on rotated testing datasets, and by comparing their respective degradations when switching from original to rotated testing data.\\\\
	More concretely, it is obvious from Table \ref{modelnet_results} how the baseline KP-CNN's performance drops by more than half its performance on the original data, while that of FA-KPCNN is nearly the same on both original and rotated data. We also notice from Table \ref{3dmatch_results_delta} positive improvements of up to $+19.7\%$ for FA-GeoTransformer vs the baseline GeoTransformer on nearly all metrics when testing on rotated data.\\
	\item \textbf{Training data scarcity}: Given that FA-KPConv models embed geometrical prior knowledge in their design, they are more training sample efficient, i.e. they need less data to learn to the same level of their KPConv counterparts, since they do not need to approximately learn the desired geometrical properties from the data. We can see this clearly in Table \ref{3dmatch_results_delta}, where the correspondence estimation improvements per testing configuration are always higher for smaller number of training samples.\\
	\item \textbf{Challenging test data}: We also notice that the prior knowledge embedded in FA-KPConv models helps with challenging test data that is not trivially related to that embedded prior knowledge. This is probably due to the fact that the learning capacity of those models can be better invested for learning more insights from the data, which enables better handling of complex scenarios. From our experiments, Table \ref{3dmatch_results_delta} shows that the improvements of FA-KPConv models for any number of training samples are more significant on 3DLoMatch than on 3DMatch.\\
\end{enumerate}
In Figure \ref{qualitative_results}, we further show some qualitative results with all three challenges being present, namely when training on 1k samples from the 3DMatch benchmark and testing on rotated 3DLoMatch data, for which the overlap between the source and target clouds is very small. We can clearly see how our method manages to robustly estimate good transformations for those very hard scenarios, while the baseline fails.\\\\
It is worth noting that we did not spend effort on tuning the hyperparameters of the models used in our experiments, since we only aimed at showing the improvement that FA-KPConv can achieve over KPConv, while using the same hyperparameters for both and the same training procedure, i.e. we focus on the relative not the absolute performance.\\\\
Also note that we did not compare our method to other methods that introduce in- or equivariance, since no other method preserves the same underlying baseline architecture and the same number of learnable parameters, thereby rendering any such comparison unfair and outside our scope of interest.
\section{Conclusion}
We introduced FA-KPConv, which can be thought of as an extension of the popular KPConv backbone, enabling the easy and flexible introduction of \textbf{exact} in- or equivariance to Euclidean transformations to KPConv-based NNs, without having to modify the amount of trainable parameters, or to alter the architecture, or to hand-craft custom input features for those NNs. We also validated the advantage of FA-KPConv on relevant benchmarks. Our method is mostly useful in the low training data regime, and when the testing data is more challenging than the data used for training.
\section*{Acknowledgment}
We thank Dr. Tobias Strau\ss{} from Robert Bosch GmbH for his valuable input, and Ms. Zeina El Kojok for her help in the implementation.

\input{fa_kpconv_paper.bbl}

\end{document}

%% file: fa_kpconv_paper.bbl

%% file: fa_kpconv_paper.bbl
\begin{thebibliography}{10}
\providecommand{\url}[1]{#1}
\csname url@samestyle\endcsname
\providecommand{\newblock}{\relax}
\providecommand{\bibinfo}[2]{#2}
\providecommand{\BIBentrySTDinterwordspacing}{\spaceskip=0pt\relax}
\providecommand{\BIBentryALTinterwordstretchfactor}{4}
\providecommand{\BIBentryALTinterwordspacing}{\spaceskip=\fontdimen2\font plus
\BIBentryALTinterwordstretchfactor\fontdimen3\font minus
  \fontdimen4\font\relax}
\providecommand{\BIBforeignlanguage}[2]{{%
\expandafter\ifx\csname l@#1\endcsname\relax
\typeout{** WARNING: IEEEtran.bst: No hyphenation pattern has been}%
\typeout{** loaded for the language `#1'. Using the pattern for}%
\typeout{** the default language instead.}%
\else
\language=\csname l@#1\endcsname
\fi
#2}}
\providecommand{\BIBdecl}{\relax}
\BIBdecl

\bibitem{kpconv}
H.~Thomas, C.~R. Qi, J.-E. Deschaud, B.~Marcotegui, F.~Goulette, and L.~J.
  Guibas, ``Kpconv: Flexible and deformable convolution for point clouds,'' in
  \emph{Proceedings of the IEEE/CVF International Conference on Computer Vision
  (ICCV)}, October 2019.

\bibitem{gfnet}
\BIBentryALTinterwordspacing
H.~Qiu, B.~Yu, and D.~Tao, ``Gfnet: Geometric flow network for 3d point cloud
  semantic segmentation,'' 2022. [Online]. Available:
  \url{https://arxiv.org/abs/2207.02605}
\BIBentrySTDinterwordspacing

\bibitem{regtr}
Z.~J. Yew and G.~H. Lee, ``Regtr: End-to-end point cloud correspondences with
  transformers,'' in \emph{Proceedings of the IEEE/CVF Conference on Computer
  Vision and Pattern Recognition (CVPR)}, June 2022, pp. 6677--6686.

\bibitem{superpoint_matching}
\BIBentryALTinterwordspacing
A.~Gupta, Y.~Xie, H.~Singh, and H.~Jiang, ``A strong baseline for point cloud
  registration via direct superpoints matching,'' 2024. [Online]. Available:
  \url{https://arxiv.org/abs/2307.01362}
\BIBentrySTDinterwordspacing

\bibitem{obpose}
Y.~Wu, O.~P. Jones, and I.~Posner, ``Obpose: Leveraging pose for object-centric
  scene inference and generation in 3d,'' \emph{arXiv preprint
  arXiv:2206.03591}, 2022.

\bibitem{geotransformer}
Z.~Qin, H.~Yu, C.~Wang, Y.~Guo, Y.~Peng, S.~Ilic, D.~Hu, and K.~Xu,
  ``Geotransformer: Fast and robust point cloud registration with geometric
  transformer,'' \emph{IEEE Transactions on Pattern Analysis and Machine
  Intelligence}, vol.~45, no.~8, pp. 9806--9821, 2023.

\bibitem{pcldl}
\BIBentryALTinterwordspacing
Q.~Zhu, L.~Fan, and N.~Weng, ``Advancements in point cloud data augmentation
  for deep learning: A survey,'' \emph{Pattern Recognition}, vol. 153, p.
  110532, 2024. [Online]. Available:
  \url{https://www.sciencedirect.com/science/article/pii/S0031320324002838}
\BIBentrySTDinterwordspacing

\bibitem{fa}
O.~Puny, M.~Atzmon, H.~Ben-Hamu, I.~Misra, A.~Grover, E.~J. Smith, and
  Y.~Lipman, ``Frame averaging for invariant and equivariant network design,''
  \emph{arXiv preprint arXiv:2110.03336}, 2021.

\bibitem{pointnet}
C.~R. Qi, H.~Su, K.~Mo, and L.~J. Guibas, ``Pointnet: Deep learning on point
  sets for 3d classification and segmentation,'' in \emph{Proceedings of the
  IEEE Conference on Computer Vision and Pattern Recognition (CVPR)}, July
  2017.

\bibitem{pointnet++}
C.~R. Qi, L.~Yi, H.~Su, and L.~J. Guibas, ``Pointnet++: Deep hierarchical
  feature learning on point sets in a metric space,'' \emph{Advances in neural
  information processing systems}, vol.~30, 2017.

\bibitem{pointnext}
G.~Qian, Y.~Li, H.~Peng, J.~Mai, H.~Hammoud, M.~Elhoseiny, and B.~Ghanem,
  ``Pointnext: Revisiting pointnet++ with improved training and scaling
  strategies,'' \emph{Advances in neural information processing systems},
  vol.~35, pp. 23\,192--23\,204, 2022.

\bibitem{spidercnn}
Y.~Xu, T.~Fan, M.~Xu, L.~Zeng, and Y.~Qiao, ``Spidercnn: Deep learning on point
  sets with parameterized convolutional filters,'' in \emph{Proceedings of the
  European Conference on Computer Vision (ECCV)}, September 2018.

\bibitem{pointcnn}
Y.~Li, R.~Bu, M.~Sun, W.~Wu, X.~Di, and B.~Chen, ``Pointcnn: Convolution on
  x-transformed points,'' \emph{Advances in neural information processing
  systems}, vol.~31, 2018.

\bibitem{paconv}
M.~Xu, R.~Ding, H.~Zhao, and X.~Qi, ``Paconv: Position adaptive convolution
  with dynamic kernel assembling on point clouds,'' in \emph{Proceedings of the
  IEEE/CVF Conference on Computer Vision and Pattern Recognition (CVPR)}, June
  2021, pp. 3173--3182.

\bibitem{pointconv}
W.~Wu, Z.~Qi, and L.~Fuxin, ``Pointconv: Deep convolutional networks on 3d
  point clouds,'' in \emph{Proceedings of the IEEE/CVF Conference on Computer
  Vision and Pattern Recognition (CVPR)}, June 2019.

\bibitem{dgcnn}
Y.~Wang, Y.~Sun, Z.~Liu, S.~E. Sarma, M.~M. Bronstein, and J.~M. Solomon,
  ``Dynamic graph cnn for learning on point clouds,'' \emph{ACM Transactions on
  Graphics (tog)}, vol.~38, no.~5, pp. 1--12, 2019.

\bibitem{ecc}
M.~Simonovsky and N.~Komodakis, ``Dynamic edge-conditioned filters in
  convolutional neural networks on graphs,'' in \emph{Proceedings of the IEEE
  Conference on Computer Vision and Pattern Recognition (CVPR)}, July 2017.

\bibitem{adaptconv}
H.~Zhou, Y.~Feng, M.~Fang, M.~Wei, J.~Qin, and T.~Lu, ``Adaptive graph
  convolution for point cloud analysis,'' in \emph{Proceedings of the IEEE/CVF
  International Conference on Computer Vision (ICCV)}, October 2021, pp.
  4965--4974.

\bibitem{pct}
M.-H. Guo, J.-X. Cai, Z.-N. Liu, T.-J. Mu, R.~R. Martin, and S.-M. Hu, ``Pct:
  Point cloud transformer,'' \emph{Computational Visual Media}, vol.~7, pp.
  187--199, 2021.

\bibitem{pt}
H.~Zhao, L.~Jiang, J.~Jia, P.~H. Torr, and V.~Koltun, ``Point transformer,'' in
  \emph{Proceedings of the IEEE/CVF International Conference on Computer Vision
  (ICCV)}, October 2021, pp. 16\,259--16\,268.

\bibitem{ptv2}
X.~Wu, Y.~Lao, L.~Jiang, X.~Liu, and H.~Zhao, ``Point transformer v2: Grouped
  vector attention and partition-based pooling,'' \emph{Advances in Neural
  Information Processing Systems}, vol.~35, pp. 33\,330--33\,342, 2022.

\bibitem{ptv3}
X.~Wu, L.~Jiang, P.-S. Wang, Z.~Liu, X.~Liu, Y.~Qiao, W.~Ouyang, T.~He, and
  H.~Zhao, ``Point transformer v3: Simpler faster stronger,'' in
  \emph{Proceedings of the IEEE/CVF Conference on Computer Vision and Pattern
  Recognition (CVPR)}, June 2024, pp. 4840--4851.

\bibitem{pointbert}
X.~Yu, L.~Tang, Y.~Rao, T.~Huang, J.~Zhou, and J.~Lu, ``Point-bert:
  Pre-training 3d point cloud transformers with masked point modeling,'' in
  \emph{Proceedings of the IEEE/CVF Conference on Computer Vision and Pattern
  Recognition (CVPR)}, June 2022, pp. 19\,313--19\,322.

\bibitem{pointmae}
Y.~Pang, W.~Wang, F.~E. Tay, W.~Liu, Y.~Tian, and L.~Yuan, ``Masked
  autoencoders for point cloud self-supervised learning,'' in \emph{European
  conference on computer vision}.\hskip 1em plus 0.5em minus 0.4em\relax
  Springer, 2022, pp. 604--621.

\bibitem{pointm2ae}
R.~Zhang, Z.~Guo, P.~Gao, R.~Fang, B.~Zhao, D.~Wang, Y.~Qiao, and H.~Li,
  ``Point-m2ae: multi-scale masked autoencoders for hierarchical point cloud
  pre-training,'' \emph{Advances in neural information processing systems},
  vol.~35, pp. 27\,061--27\,074, 2022.

\bibitem{stn}
M.~Jaderberg, K.~Simonyan, A.~Zisserman \emph{et~al.}, ``Spatial transformer
  networks,'' \emph{Advances in neural information processing systems},
  vol.~28, 2015.

\bibitem{gipointnet}
H.~Le, ``Geometric invariance of pointnet,'' {B.S.} thesis, 2021.

\bibitem{gecnn}
T.~Cohen and M.~Welling, ``Group equivariant convolutional networks,'' in
  \emph{International conference on machine learning}.\hskip 1em plus 0.5em
  minus 0.4em\relax PMLR, 2016, pp. 2990--2999.

\bibitem{stcnns}
T.~S. Cohen and M.~Welling, ``Steerable cnns,'' \emph{arXiv preprint
  arXiv:1612.08498}, 2016.

\bibitem{spcnns}
T.~S. Cohen, M.~Geiger, J.~K{\"o}hler, and M.~Welling, ``Spherical cnns,''
  \emph{arXiv preprint arXiv:1801.10130}, 2018.

\bibitem{epn}
H.~Chen, S.~Liu, W.~Chen, H.~Li, and R.~Hill, ``Equivariant point network for
  3d point cloud analysis,'' in \emph{Proceedings of the IEEE/CVF conference on
  computer vision and pattern recognition}, 2021, pp. 14\,514--14\,523.

\bibitem{tfn}
N.~Thomas, T.~Smidt, S.~Kearnes, L.~Yang, L.~Li, K.~Kohlhoff, and P.~Riley,
  ``Tensor field networks: Rotation-and translation-equivariant neural networks
  for 3d point clouds,'' \emph{arXiv preprint arXiv:1802.08219}, 2018.

\bibitem{se3tr}
F.~Fuchs, D.~Worrall, V.~Fischer, and M.~Welling, ``Se (3)-transformers: 3d
  roto-translation equivariant attention networks,'' \emph{Advances in neural
  information processing systems}, vol.~33, pp. 1970--1981, 2020.

\bibitem{clusternet}
C.~Chen, G.~Li, R.~Xu, T.~Chen, M.~Wang, and L.~Lin, ``Clusternet: Deep
  hierarchical cluster network with rigorously rotation-invariant
  representation for point cloud analysis,'' in \emph{Proceedings of the
  IEEE/CVF conference on computer vision and pattern recognition}, 2019, pp.
  4994--5002.

\bibitem{riconv}
Z.~Zhang, B.-S. Hua, D.~W. Rosen, and S.-K. Yeung, ``Rotation invariant
  convolutions for 3d point clouds deep learning,'' in \emph{2019 International
  conference on 3d vision (3DV)}.\hskip 1em plus 0.5em minus 0.4em\relax IEEE,
  2019, pp. 204--213.

\bibitem{ppfnet}
H.~Deng, T.~Birdal, and S.~Ilic, ``Ppfnet: Global context aware local features
  for robust 3d point matching,'' in \emph{Proceedings of the IEEE conference
  on computer vision and pattern recognition}, 2018, pp. 195--205.

\bibitem{srinet}
X.~Sun, Z.~Lian, and J.~Xiao, ``Srinet: Learning strictly rotation-invariant
  representations for point cloud classification and segmentation,'' in
  \emph{Proceedings of the 27th ACM international conference on multimedia},
  2019, pp. 980--988.

\bibitem{darboux}
F.~P. Ferrie, J.~Lagarde, and P.~Whaite, ``Darboux frames, snakes, and
  super-quadrics: Geometry from the bottom up,'' \emph{IEEE transactions on
  pattern analysis and machine intelligence}, vol.~15, no.~8, pp. 771--784,
  1993.

\bibitem{egnn}
V.~G. Satorras, E.~Hoogeboom, and M.~Welling, ``E (n) equivariant graph neural
  networks,'' in \emph{International conference on machine learning}.\hskip 1em
  plus 0.5em minus 0.4em\relax PMLR, 2021, pp. 9323--9332.

\bibitem{vn}
C.~Deng, O.~Litany, Y.~Duan, A.~Poulenard, A.~Tagliasacchi, and L.~J. Guibas,
  ``Vector neurons: A general framework for so (3)-equivariant networks,'' in
  \emph{Proceedings of the IEEE/CVF International Conference on Computer
  Vision}, 2021, pp. 12\,200--12\,209.

\bibitem{rathsurvey}
M.~Rath and A.~P. Condurache, ``Boosting deep neural networks with geometrical
  prior knowledge: A survey,'' \emph{Artificial Intelligence Review}, vol.~57,
  no.~4, p.~95, 2024.

\bibitem{modelnet}
Z.~Wu, S.~Song, A.~Khosla, F.~Yu, L.~Zhang, X.~Tang, and J.~Xiao, ``3d
  shapenets: A deep representation for volumetric shapes,'' in
  \emph{Proceedings of the IEEE conference on computer vision and pattern
  recognition}, 2015, pp. 1912--1920.

\bibitem{3dmatch}
A.~Zeng, S.~Song, M.~Nie{\ss}ner, M.~Fisher, J.~Xiao, and T.~Funkhouser,
  ``3dmatch: Learning local geometric descriptors from rgb-d reconstructions,''
  in \emph{Proceedings of the IEEE conference on computer vision and pattern
  recognition}, 2017, pp. 1802--1811.

\end{thebibliography}
